\documentclass[times,final]{article}

\usepackage[inline, draft]{fixme}

\fxsetup{theme=color, mode=multiuser} \FXRegisterAuthor{me}{ame}{\color{red}Me}
\usepackage{framed,multirow}
\usepackage{acronym}
\usepackage{booktabs}
\usepackage{graphicx}
\usepackage{subcaption}
\usepackage{multicol}
\usepackage{amssymb}
\usepackage{latexsym}
\usepackage{authblk}
\usepackage{natbib}
\setcitestyle{authoryear,open={(},close={)}}
\usepackage{url}
\usepackage{xcolor}
\usepackage{blindtext}
\usepackage[]{algorithm2e}
\usepackage{amsmath}
\usepackage{hyperref}
\usepackage{float}
\DeclareMathOperator*{\argmax}{arg\,max}
\DeclareMathOperator*{\argmin}{arg\,min}

\usepackage{hyperref}
\definecolor{newcolor}{rgb}{.8,.349,.1}

\usepackage{fancyhdr}
\begin{document}

\title{Handling confounding variables in statistical shape analysis - application to cardiac remodelling}
\author[1]{Gabriel Bernardino}

\author[2]{Oualid {Benkarim}}
\author[3, 4]{Mar\'ia {Sanz-de la Garza}}

\author[3,4]{Susanna {Prat-Gonz\`alez}}
\author[6,7]{Alvaro {Sepulveda-Martinez}}
\author[4,6,8]{F\`atima {Crispi}}
\author[3,4,5]{Marta {Sitges}}
\author[9]{Constantine {Butakoff}}
\author[10]{Mathieu {De Craene}}
\author[1,4,11]{Bart {Bijnens}}
\author[1,11]{Miguel A. {Gonz\'alez Ballester}}


\affil[1]{BCN Medtech, Dept. of Information and Communication Technologies, Universitat Pompeu Fabra,  Barcelona, Spain}
\affil[2]{McConnell Brain Imaging Centre, Montreal Neurological Institute and Hospital, McGill University, Montreal, Canada}
\affil[3]{Cardiovascular Institute, Hospital Clínic, Barcelona, Spain}
\affil[4]{IDIBAPS, Barcelona, Spain}
\affil[6]{BCNatal, Hospitals Clinic and Sant Joan de Déu, Universitat de Barcelona, Barcelona, Spain}
\affil[7]{Fetal Medicine Unit, Department of Obstetrics and Gynecology, Hospital Clínico de la Universidad de Chile. Santiago de Chile, Chile}
\affil[8]{CIBER-ER, Barcelona, Spain}
\affil[5]{CIBERCV, Barcelona, Spain}

\affil[9]{Barcelona Supercomputing Center , Barcelona, Spain}
\affil[10]{Philips Medisys, Paris, France}
\affil[11]{ICREA, Barcelona, Spain}
\maketitle

\newacro{SSA}{Statistical shape analysis}
\newacro{MRI}{ magnetic resonance imaging}
\newacro{SA}{ short axis}

\newacro{PLS}{ partial least squares}
\newacro{PCA}{ principal components analysis}
\newacro{PGA}{ principal geodesic analysis}
\newacro{GPPA}{ generalized partial procrustes analysis}

\newacro{PDM}{ point distribution model}
\newacro{CV}{ cross validation}

\newacro{DR}{ dimensionality reduction}

\newacro{BMI}{body mass index}
\newacro{BSA}{body surface area}

\newacro{LV}{left ventricle}
\newacro{RV}{right ventricle}
\newacro{ED}{end diastolic}
\newacro{ES}{end systolic}

\newacro{CO}{cardiac output}

\newpage
\begin{abstract}

Statistical shape analysis is a powerful tool to assess organ morphologies and find shape changes associated to a particular disease. However, imbalance in confounding factors, such as demographics might invalidate the analysis if not taken into consideration. Despite the methodological advances in the field, providing new methods that are able to capture complex and regional shape differences, the relationship between non-imaging information and shape variability has been overlooked.

We present a linear statistical shape analysis framework that finds shape differences unassociated to a controlled set of confounding variables. It includes two confounding correction methods: confounding deflation and adjustment. We applied our framework to a cardiac magnetic resonance imaging dataset, consisting of the cardiac ventricles of 89 triathletes and 77 controls, to identify cardiac remodelling due to the practice of endurance exercise. To test robustness to confounders, subsets of this dataset were generated by randomly removing controls with low body mass index, thus introducing imbalance.

The analysis of the whole dataset indicates an increase of ventricular volumes and myocardial mass in athletes, which is consistent with the clinical literature. However, when confounders are not taken into consideration no increase of myocardial mass is found. Using the downsampled datasets, we find that confounder adjustment methods are needed to find the real remodelling patterns in imbalanced datasets.
\end{abstract}

\section{Introduction}
Analysing the shapes of parts of biological organs and organisms has been the object of extensive study for over a century  \citep{Thompson1942}. This interest in shape is also present in medicine: several studies have focused in the relationship between organ morphology and illness. For instance, cardiac shape remodels to improve cardiac pressure/volume output under abnormal working conditions, and it is used to assess the presence/evolution of illness \citep{Arts1994,Grossman1975}. In a nutshell, pressure overload produces concentric remodelling (thickening of the myocardium without dilation of the ventricle) to maintain wall stresses low, and volume overload dilates the ventricle without a myocardial mass thickening. This is an oversimplication, as a volume overload will also increase pressure, and the exact remodelling mechanisms and triggers at a cellular level are still under research and discussion.  The classical way of analysing shape in the clinical community consists in manually extracting hand-crafted features, and analysing these shape descriptors. These measurements are usually standardised and defined in guidelines \citep{Lang2015}, and usually refer to global characteristics of the shape  that carry little regional information. 
 
Nowadays, it is possible to acquire 3D images in clinical routine. Furthermore, advances in computing permit to automatically segment the images and to generate personalised 3D models of the organs \citep{Ballester2000,Mitchell2002,Ecabert2006,Bernard2018}. \ac{SSA}  is a set of techniques to  represent both shapes and images and do the analysis directly with these objects, not being limited to only analyse previously defined measurements.  \ac{SSA}  is used in the medical imaging field, in order to identify and represent shape variability of those organs \citep{Cerrolaza2015,Blanc2012,Rajamani2007,Sierra2006}.  This allows expressing and quantifying regional shape patterns in a robust and objective manner, instead of working with a small set of predefined measurements on the shapes (like volumes and diameters). \ac{SSA} can be used together with statistical learning techniques to construct models that find regional differences in anatomy that are associated to pathologies \citep{Zhang2014, Singh2014a, Varano2017, Sarvari2017}, based on a control and a pathological population. Roughly, the typical framework consists of first using \ac{SSA} to construct an atlas of all shapes in the population, then  use \ac{PCA} or another \ac{DR} technique  to find a low-rank representation of the shape space, and finally use a classification algorithm on that space to train a model that predicts the control/pathologic status. 

Beyond pathological variability, the shape of an organ also exhibits variability due to other factors, like lifestyle, gender, ethnicity, or size. The framework described above uses the implicit hypothesis that differences in shape are only due to the pathology.  In some cases the remodelling associated to the pathology is prominent and easily identifiable. However, in others, like subclinical  or  early-stage studies, differences  can often be very subtle and less pronounced than demographic-related variability. \textcolor{black}{Given the difficulty to acquire medical data, many imaging studies are cross-sectional and observational, and participants are recruited prospectively. As imaging datasets have typically low sample sizes, it is not always possible to obtain balanced subdatasets, and researchers are forced to analyse imbalaced datasets. If left uncorrected this imbalance in the population characteristics may result in wrong conclusions/models \citep{Cohen2018DistributionTranslation}}. Even in cases where the pathological remodelling is significant, and the populations are similar in terms of demographics,  demographic-related variability will add noise to the analysis. A number of authors have explored the usage of non-imaging information. For instance  Singh et al. proposed a procedure similar to partial correlation between shape and several clinical variables while correcting for confounding variables\citep{Singh2014a}. Zhang et al\textcolor{black}{.} adjusted by demographics in their studies of cardiac remodelling in myocardial infarction \citep{Zhang2014a}, and Zhang et al\textcolor{black}{.} and Mauger at al\textcolor{black}{.}  explored the relationship between shape and classical clinical measurements \citep{Zhang2017, Zhang2014a,Mauger2019}. However, not all authors include corrections for confounders, and their effect in shape analysis  studies has not been yet quantitatively tested.

In this paper, we present a \ac{SSA} framework to find differences between control and pathological populations that outputs the most discriminating shape pattern, that can be visualised for interpretability.  We quantitatively and qualitatively show the effect an imbalance  in the confounding variables has in the analysis, and propose techniques to reduce that effect. The proposed model consists of the following steps: (1) the construction of an atlas of the personalised 3D meshes automatically generated from images; (2) identification and removal of shape variability due to confounding variables;  (3) dimensionality reduction and classification.

To illustrate the framework, we use a dataset of cardiac \ac{MRI} involving sedentary controls and triathlon athletes. This dataset was  collected to study the remodelling due to the extended practice of endurance sport, which produces a volume overload to the heart. This volume overload triggers compensatory mechanisms to improve cardiac output and withstand the increased pressure during exercise. The whole of this remodelling is called the \emph{Athlete's Heart} and involves substantial changes in function and geometry at both rest and during exercise  \citep{DAndrea2015, Schiros2013a}. Although the remodelling is not yet completely understood, researchers have established a strong relationship between cardiopulmonary performance during exercise and cardiac geometry at rest \citep{LaGerche2012,  Scharhag2002}.

\section{Methodology}

\begin{figure*}[!t]
\centering
\includegraphics[width = .9\textwidth]{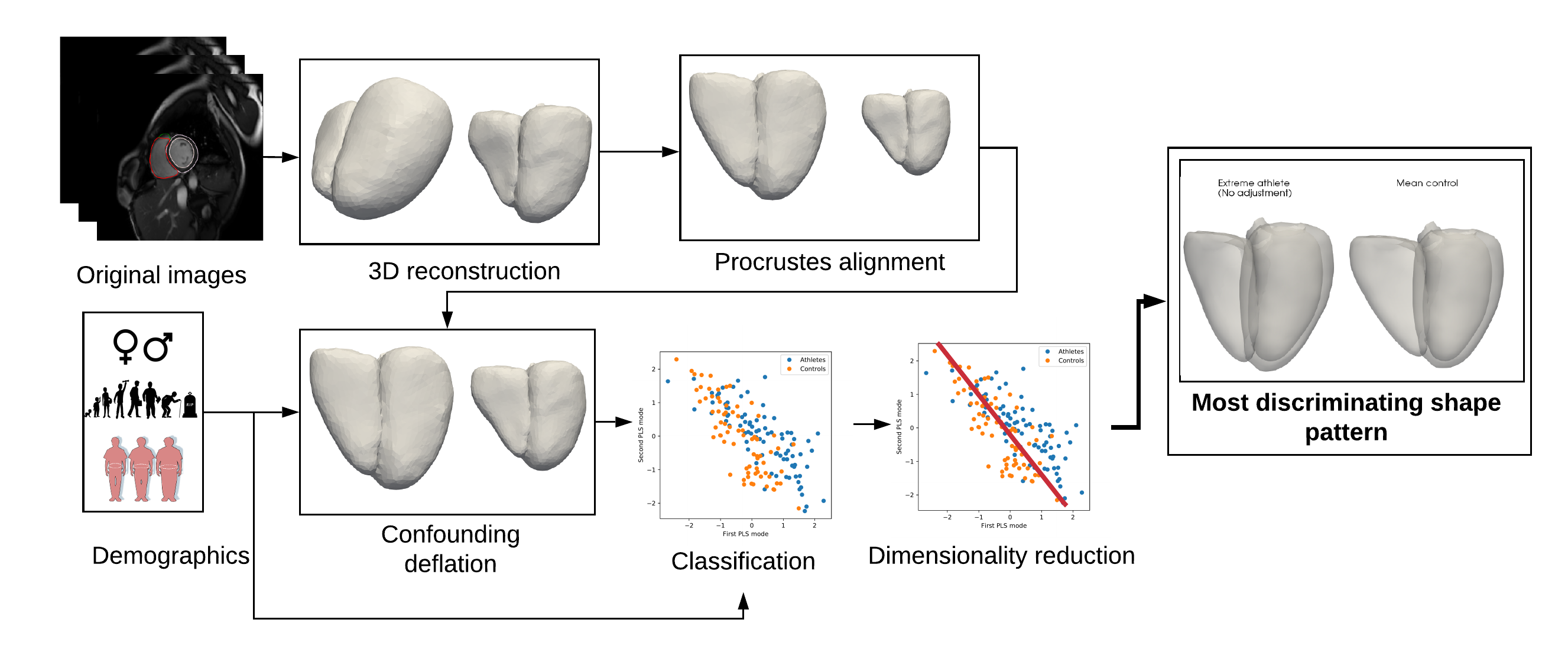}
\caption[Schema of the SSA framework]{Schema of the framework and its different components. The input are the short axis MRI, and the demographics of the population. From the image, we generate a personalised 3D mesh of the ventricles, and align them using Procrustes analysis. Afterwards, we remove confounding-related shape variability using confounding deflation. This is followed by a dimensionality reduction and a classification steps. The final step is to compute the most discriminative shape pattern from the classification model coefficients.}
\label{methodology:figure:framework}
\end{figure*}

The full process to compute the confounder invariant most discriminating shape pattern between two populations is summarised in Figure \ref{methodology:figure:framework}. The presented framework consists of $3$ main steps: 
\begin{enumerate}
\item Compute the mean shape of the population, and register all shapes to this template.
\item Identify and remove shape variability attributable to confounding variables.
\item Train a classification model to obtain the most discriminating shape pattern between both populations and generate a visual representation of the most discriminating shape pattern
\end{enumerate}

\subsection{Atlas construction}
From the \ac{SA} \ac{MRI} sequence, we use a model-based  automatic segmentation method to obtain personalised meshes with point-to-point correspondence \citep{Ecabert2006}.  The method deforms a full heart (4 chambers) template mesh using a polyaffine deformation to match the myocardial boundaries. The method includes slice correction to remove misalignment between consecutive  slices.

Since only the ventricles were visible in our images, we discarded the atria and big vessels from each segmented shape. Each resulting mesh has 4446 vertices (the \ac{LV} has 3052 vertices and the \ac{RV} has 1776 vertices, the right-most part of the septum belongs to both ventricles) in point-to-point correspondence, and 9004 triangles.  Only the \ac{ED} frame is selected for the analysis. Since meshes are in point-to-point correspondence and share the same connectivity, they can be analysed using the \ac{PDM} \citep{Cootes1995a}. In \ac{PDM}, the shape of each patient $j$ is associated to a vector with the concatenated position $(x, y, z)$ of its nodes, giving a shape vector:
\begin{equation}
X_j = ( x_0, y_0, z_0, x_1, y_1, z_1 \dots x_N, y_N, z_N)
\end{equation}

\textcolor{black}{A problem with this parametrisation is that it does not avoid degenerate self-intersecting shapes. More complex techniques have been developed, in which meshes are represented by the action of diffeormophism to a given reference \citep{Ashburner2011DiffeomorphicOptimisation,Miller2006GeodesicAnatomy,Durrleman2014}. These representations avoid self-intersections and provide a more accurate shape representation, but have a bigger computational cost. We chose \ac{PDM} over these representations since we have not observed any of the non-diffeomorphic artefacts. In supplementary material S1 we have compared the metric induced by these methods with the \ac{PDM} and found only minor differences.} We applied generalised Partial Procrustes Analysis \citep{Dryden1998a} to align the meshes thus removing the positioning and orientation variability. We maintained size during this step, but differences in size due to anthropometric variables will be identified and removed in other steps of the framework. This algorithm is an iterative method, at each iteration computes an estimation of the mean shape $\bar{X}^j$  and then rigidly registers each shape to that estimated mean. This is repeated until convergence, in order to obtain an unbiased mean. Here we show a full iteration $j$ of the algorithm:
\begin{equation}
\bar{X}^j = \frac{1}{n} \sum^n X_i^j
\end{equation}
\begin{equation}
\forall i: \; R_i^j, t_i^j = \argmin_{R, t} \|R (X_i - t) - \bar{X}^j \|^2
\end{equation}
\begin{equation}
X_i^{j + 1} = R_i^j(X_i - t_i)
\end{equation}


\subsection{Confounding deflation}
\label{methodology:subsection:residual}
To identify and remove the  shape variability related to the confounding variables ($M$), and not to the studied condition, we use a procedure similar to partial correlation, as done in \citep{Singh2014a} for a regression problem. We assume that shape $X$ can be decomposed in the sum of the population mean ($\mu_X$) and some deformation from that mean that is composed of noise ($\epsilon$), variability caused by the confounding variables ($X_M$) and variability from other sources ($X_i$). The last component includes remodelling due to pathology :
\begin{equation}
X = \mu_X + X_M + X_i +  \epsilon
\end{equation}

The first step consists in estimating $X_M$ by building a linear model (whose coefficients are $w_M$) that predicts the expected shape from the confounding variables $M$. The training data of this model is only one population (the controls), to avoid introducing possible inter-population differences in the model. This prediction will be our $X_M$. Then, for the shape of each individual, we subtract the predicted shape from the actual shape obtaining the prediction residuals.  Residuals represent the part of the shape variability that cannot be explained by the confounding variables.  To maintain the residual vectors in the same range as the original shape vectors, we summate the original population shape mean. The final formula reads:

\begin{equation}
X_{res} =  X - X_M =  (X - w_c\cdot M) +\mu_X 
\end{equation}

The regression model coefficients $w_c$ are computed using \ac{PLS} with the shapes $X$ and confounding variables $M$. \ac{PLS} is a regression method that projects the input and output data to two low-dimension subspaces (called embeddings)  that have maximal covariance  \citep{Wegelin2000}. The confounding variables are standardised to have $1$ standard deviation and $0$ mean, but  shapes are only standardised to have $0$ mean. The full process is described below, and consists in an iterative process where at each iteration a new regression dimension of the low-dimensional spaces is computed. The predicted part is then removed from the input and output spaces, and the process is repeated until the desired number of iterations is reached. Several versions of \ac{PLS} exist, and we used Wold's version. All \ac{PLS} versions agree in the first iteration, but give different result for embedding spaces of more than one dimension.  Here we show the algorithm for the $r$-th iteration of a \ac{PLS} associating an input space $X$ with some response $Y$, both being matrices.  In our setting, and only for this part $X$ will be the confounding variables ($M$ in the rest of the paper) and $Y$ the shape vectors ($X$ in the rest of the paper).

First, we compute the new dimensions of the embedding spaces by solving an eigenvector problem:
\begin{equation}
\label{method:eq:dimRedPLS}
u^r, v^r  = \argmax_{\|u^*\| = 1,  \|v^*\| = 1} {{u^*}^t X^t Y v^*}
\end{equation}

We compute the rank one approximations (hat variables), and the regression coefficient $w_r$; for OLS refers to the classical ordinary least squares.
\begin{equation}
\begin{split}
\hat{X}^r = OLS_{prediction}(X^r u^r, Y^r v^r) \\
w^r = v^r (u^r)^t  \frac{(X^r u^r)^t Y^r v^r }{X^r u^r,} \\
 \hat{Y}^r = OLS_{prediction}(Y^r u^r, X^r v^r )
\end{split}
\end{equation}

These rank one optimisations are used to update the input $X$ and its output $Y$ by removing the part that is already predicted.
\begin{equation}
\begin{split}
X^{r + 1} = X^r - \hat{X}^r  \\
Y^{r + 1} = Y^r - \hat{Y}^r \end{split}
\end{equation}

The final result of the algorithm is the regression coefficients $w$ from $X$ to $Y$; and it is obtained by summing all the one-rank approximation regression coefficients.
\begin{equation}
w = \sum_r w^r 
\end{equation}

The coefficients $w_i$ of the prediction model, associated to confounder variable $M^i$, of this model can also be visualised and interpreted. We can consider the partial determination coefficient, and visualise the shape pattern most associated to a certain confounder, as if it were a discriminating shape (see subsection \ref{methodology:subsection:classification}).

\subsubsection{Dimensionality reduction}
Given the high dimensionality of the shape vectors and low number of samples, we use a \ac{DR} method to find a subspace that contains the most relevant shape patterns. \textcolor{black}{We choose linear methods over non-linear alternatives due to their higher interpretability, as the projection function is computed exactly. Additionally, an initial exploration of a non-linear appraoch (see supplementary material S2) did not lead to different results} The general linear \ac{DR} model reads:

\begin{equation}
 X_{orig} = \mu_X  + K X_{red} + \epsilon 
\end{equation}

Each shape $X_{orig}$ is expressed as the population mean($\mu_x$) plus some shape-specific associated low-dimensionality vector $X_{red}$ and a noise term $\epsilon$. The embedding matrix $K$ is constructed depending on the dimensionality reduction method, and contains the most interesting shape directions according to a certain metric. 

We use three different methods of linear \ac{DR}: \ac{PCA}, \ac{PLS} and a combination of both. \ac{PCA} and \ac{PLS} have been reported to be used in conjunction to classification methods, and in particular logistic regression \citep{Bastien2005}. We have described the regression modality of  \ac{PLS} in subsection \ref{methodology:subsection:residual}, but \ac{PLS} also computes a \ac{DR} space. The difference is that we discard the regression coefficients $w$ and only use embedding vectors  $u^r$ in equation \ref{method:eq:dimRedPLS} of the input space. These vectors define a vector subspace, but are not guaranteed to be orthogonal, so we use $QR$ decomposition \citep{Lord2007} to obtain an orthonormal base. The combined method is based on prefiltering the shape using \ac{PCA}, keeping a  high number of components ($>90\%$ of the variance), to then use as input to a \ac{PLS}. This decreases the computation time, and adds stability by denoising the data. Contrary to the typical procedure in machine learning, we chose not to standardise the \ac{PCA} modes by variance before applying \ac{PLS}, as the variance of each \ac{PCA} mode carries important information of the signal-to-noise-ratio.

\subsection{Classification}
\label{methodology:subsection:classification}
In this section, we present the method used to train the classifier model and use the model coefficients to obtain the most discriminating shape. The shape features obtained from the \ac{DR} are combined with the confounding variables in a logistic regression model. We choose logistic regression because we expect not to have a complete separation between both populations, and we want the model to be simple and interpretable. The logistic model gives a probability that an individual $j$ with shape $X_j$ and confounding variables $M_j$ belongs to the pathological or control populations. 
\begin{equation}
Pr(y_j = \textit{control}\ | X_j, M_j) = logit(\langle X_j, w_X \rangle + \langle M_j  w_M \rangle + b) 
\end{equation}

Logit refers to the logistic function $x \mapsto 1/(1 + exp(-x) )$. $w_X$ and $w_M$  are the logistic regression coefficients for the shape and confounding variables respectively, and are chosen to  minimize the log-loss of the probability of the training data $X$. The log-loss is the logarithm of the probability that the model  is inconsistent with the observed data:

\begin{equation}
\label{eq:logloss}
\frac{1}{n}\sum^n \log \left(\left| y_j - Pr(y_j = 1 | X_j, M_j) \right|\right)
\end{equation}

The logistic regression coefficients associated to the shape $w_X$, can be mapped from the reduced shape space to the full space by using the pseudoinverse of the dimensionality reduction matrix. Let $w_{X_{red}}$ be the coefficients associated to the reduced shape models, and $K_{PCA}$ and $K_{PLS}$ the projection matrices of \ac{PCA} and \ac{PLS} respectively. Then the coefficients associated with the full shape are:
\begin{equation}
w_X = K_{PCA}^t K_{PLS}^t w_{X_{red}}
\end{equation}

 Then, we can visualise and interpret the shape pattern.  Analogously  to multivariate regression, where  the coefficients are indexed by the standard deviation to allow comparison among them, we need to adjust for differences in variance of the different coordinates. Since node coordinates carry no meaning on their own, we treat shape as an object itself and do \ac{PCA} whitening of the shape, as is typically done with other multidimensional signals \citep{Kessy2018}. Since we are only interested in the remodelling direction, we normalise the vector to be unitary in the $L_2$ norm. With these corrections, all the shape features are correctly scaled by their importance in prediction. The full process to find the standardised shape pattern $\hat{w}$ reads:

\begin{equation}
\hat{w} =\frac{ \Sigma^{1/2} w_X}{ \| \Sigma^{1/2} w_X \|}
\end{equation}

Where $\Sigma$ is the covariance matrix estimated using \ac{PCA}. For visualisation, we can generate shapes that are representative of that shape pattern by adding the remodelling shape pattern, scaled with a parameter $\lambda$, to the mean shape. To keep the shapes within the original range, we impose that $\lambda$ has to be within $3$ standard deviations of the variance associated to the shape pattern.

\begin{equation}
\label{method:eq:syntheticMeshes}
X_{repr} (\lambda) = \mu_{X} +  \lambda \hat{w}  
\end{equation}
 
We can quantify the presence of remodelling in each shape, obtaining a scalar score for each individual, by computing the dot product of the shape vectors with the raw logistic regression coefficients associated to the shape only. The previous \ac{PCA} whitening is only done for visualisation and comparison of modes. If the shape pattern needs to be quantified in a population, the original one without standardization needs to be used. 

\begin{equation}
\textit{score}_i = \langle w_X, X_i - \mu_X  \rangle
\end{equation}

The shape patterns can be compared using the $L_2$ dot product between standardised shape vectors, which coincides with the correlation of the scores associated to each pattern.

\section{Experimental setup}
\subsection{Clinical dataset}

The study comprises 77 controls and 89 athletes that underwent a \ac{MRI}, to study the cardiac remodelling triggered by the practice of endurance sport. The study was approved by a local ethical board, and all participants gave written informed consent for the handling of their data. Recruited athletes had been training an endurance sport, triathlon, over 10h a week during the last 5 years. \textcolor{black}{All the study participants were Caucasian and none  had previous cardiovascular disease, nor were any detected during the study. More details on the recruitment protocol are found in \cite{Bernardino}.} Table \ref{clinical:demographics} shows the demographics of both populations. 

The controls and athletes come from different studies, and the demographics of both populations did not match exactly in age, but roughly represent the same general population in age and gender. The study protocol and radiologist were the same for both cases. Age is statistically different between athletes and controls, but the difference is very small (2 years). We do not expect big differences due to this imbalance, since both athletes and controls are middle-aged. There are also statistically significant differences in both weight and \ac{BSA}, but these correspond to physiological remodelling since endurance athletes are obviously fitter than  the general population. We used as possible confounders age, \ac{BSA} and gender, the typical adjustment variables in cardiology studies.

The \ac{MRI} acquisition was ECG-gated from the R-peak during breath-hold. The \ac{MRI} machines were Siemens Aereo and Siemens Magnetom, with an in-plane spatial resolution ranging from 0.5mm to 1mm.  The spacing between slices range between 8mm and 10.4mm, and the slice thickness was 8mm. \ac{MRI} sequences were acquired with 25 frames per cardiac cycle.  The ventricular contours (epicardium and endocardium in the case of the \ac{LV}, and only epicardium of the \ac{RV} ) were automatically segmented from the \ac{MRI} SA using the automatic procedure described in the methodology. \textcolor{black}{ED images were selected as the ones with maximal \ac{LV} volume, and the rest of the cardiac cycle was discarded.}

The quality of the automatic segmentations was assessed by one of our experts, but no manual refinement was performed in order to preserve point-wise correspondences.  Cases where errors could not be considered to be small were discarded: two individuals (both of them athletes) were discarded because the segmentation was inconsistent with the image. The segmentation and registration errors are handled as noise in our study. The meshes were very uniform and we found no self-intersecting artefacts. As a consequence of the thick slices, the apex was not correctly imaged and presented much more noise than the basal part of the ventricles. \textcolor{black}{No extra processing was done to correct for potential apex artefacts. We decided to leave the apex in the study to avoid boundary effects.}
\begin{table*}[!t]
\caption[ Population demographics of the study participants. Athletes have a lower heart rate and weight than controls]{\label{clinical:demographics} Population demographics of the study participants. Athletes have a lower heart rate and weight than controls. The age is significantly different, but both cohorts are middle-aged and we do not expect major age-related differences. The p-values are obtained using a Mann-Whitney test. }

\centering
\begin{tabular}{llll}
\toprule
& Athletes              & Controls   & p-value                     \\
\midrule
Age {[}y{]}           & 35.4(6.1)  & 33.4(3.8)  & 0.013           \\
BSA {[}m$^2${]}          & 1.78(0.19) & 1.86(0.20) & 0.005           \\
Weight {[}kg{]}       & 66.8(11.3) & 73.5(15.1) & 0.001           \\
Height {[}m{]}        & 1.71(0.09) & 1.73(0.08) & 0.151           \\
Women {[}\%{]}        & 0.48       & 0.44       & 0.938           \\
Resting HR  {[}bpm{]} & 57.2(8.4)  & 65.8(10.6) & \textless 0.001 \\
\bottomrule
\end{tabular}
\end{table*}

 \subsection{Automatic measurements}
We computed automatic measurements of the 3D shapes that are analogous to the classical clinical measurements, using the point-to-point correspondence and labelling coming from the model-based registration and segmentation. This allowed a better understanding of the discriminative shape patterns by assessing how these measurements vary in response to the remodelling score $\lambda$ on the synthetically generated meshes according to equation \ref{method:eq:syntheticMeshes}. We computed the \ac{ED} volumes of both ventricles, as well as the myocardial mass of the \ac{LV}.

\subsection{BMI-based downsampling}
 Obesity (defined as \ac{BMI} > 30)  and overweight (defined as \ac{BMI} > 25) have been reported as risk factors to cardiovascular illnesses in the literature  and have a clear influence in cardiac shape and function \citep{Alpert2018}. Surprisingly, overweight and  athletic remodelling share  similarities. Even if one might expect them to be opposite, as endurance athletes and overweight body fat are in the opposite sides of the spectrum, both remodellings are triggered by an increase of the heart's loading. In the case of the athletes, it is the increase of \ac{CO} during exercise that produces a volume overload, and in the obese it is a mix of increased \ac{CO} needs at rest to account for the bigger body size \citep{Lavie2007} and a ventricular pressure overload due to an increase of arterial pressure \citep{Messerli1982}. 

To study the effect of an imbalance in a shape-affecting variable between the control and case populations, we biased our population to increase the \ac{BMI} of the controls, as overweight has a well known effect on the heart. To generate the imbalance, the control class was downsampled to maintain only 25\% of its controls, favouring keeping the ones with higher \ac{BMI}. The individuals to remove were selected randomly among the controls, with a probability of being kept proportional to the rank of its \ac{BMI} in the control population. Athletes were not downsampled. This procedure was repeated with 100 different seeds to obtain different imbalanced datasets and add robustness to the results.  

We analysed how this imbalance affected the remodelling pattern found, and to which extent could confounding adjustment and confounding deflation correct this effect. To study the stability of the shape pattern, we  computed the $L_2$ product between the discriminative shape patterns obtained with the downsampled datasets and the results obtained with the full dataset, that serves as groundtruth. This was performed for the different \ac{DR} methods. We also tested the $L_2$ product of the discriminative patterns with the \ac{BMI}-shape pattern, obtained with an adaptation of our framework to the regression problem. Additionally, we compared qualitatively the automatic measurement response to the remodelling score of both the downsampled and original dataset. We tested both the covariate adjustment, and the confounding deflation. For the confounding deflation, we also evaluated how different choices during the training of the shape prediction model affected the obtained most discriminative shape pattern.

\section{Results}
\subsection{Dimensionality reduction}
To choose the best configuration of parameters for the \ac{DR} method, we used 10-fold \ac{CV} and computed the mean log-loss of all the validation set, defined in equation \ref{eq:logloss}, over a wide combination of parameters. In table \ref{experiments:table:cvTableReduced} we find the log-loss of the best parameter choice for the 3 different \ac{DR} methods  (PCA, PLS and PCA + PLS). They correspond to a PCA with 5 modes, a PLS with 3 and the PCA + PLS with 20 \ac{PCA} modes and 3 \ac{PLS} modes. \textcolor{black}{In the supplementary material S3, Table 3  shows the results of all the parameter combinations tested.} This experiment also provided an overview of how the use of demographics affected the classification metrics: there is a considerable improvement in terms of log-loss when the confounding variables are used in the model, and a minor improvement when using PLS instead of PCA. When using PCA + PLS, the metric is very similar to PLS. Confounding deflation gave a worse result than the raw (non-deflated) shapes when adjustment is used. However, confounding deflation improved the resulting classification metric with respect to the raw when confounders are not added to the logistic regression model.
\begin{table*}[!h]
\caption[Results of DR model selection]{10-fold CV log-loss scores of the best choice for each \ac{DR} method.}
\label{experiments:table:cvTableReduced} 
\centering
\begin{tabular}{lrrrr}
\toprule
{} &  \multicolumn{2}{c}{No deflation} & \multicolumn{2}{c}{No deflation} \\

Method &  No adj. &  Confounders adj. &  No adj. &  Confounders adj. \\
\midrule
PCA$_5$     &                  0.58 &                          0.46 &                  0.48 &                          0.46 \\
PLS$_3$     &                  0.59 &                          0.44 &                  0.52 &                          0.48 \\
PCA$_{20}$ + PLS$_3$ &                  0.59 &                          0.43 &                  0.50 &                          0.46 \\
\bottomrule
\end{tabular}T
\end{table*}

\subsection{Athletic model }
\begin{figure*}[!t]
\begin{subfigure}[b]{0.4\textwidth}
\centering
\includegraphics[width = .9\textwidth]{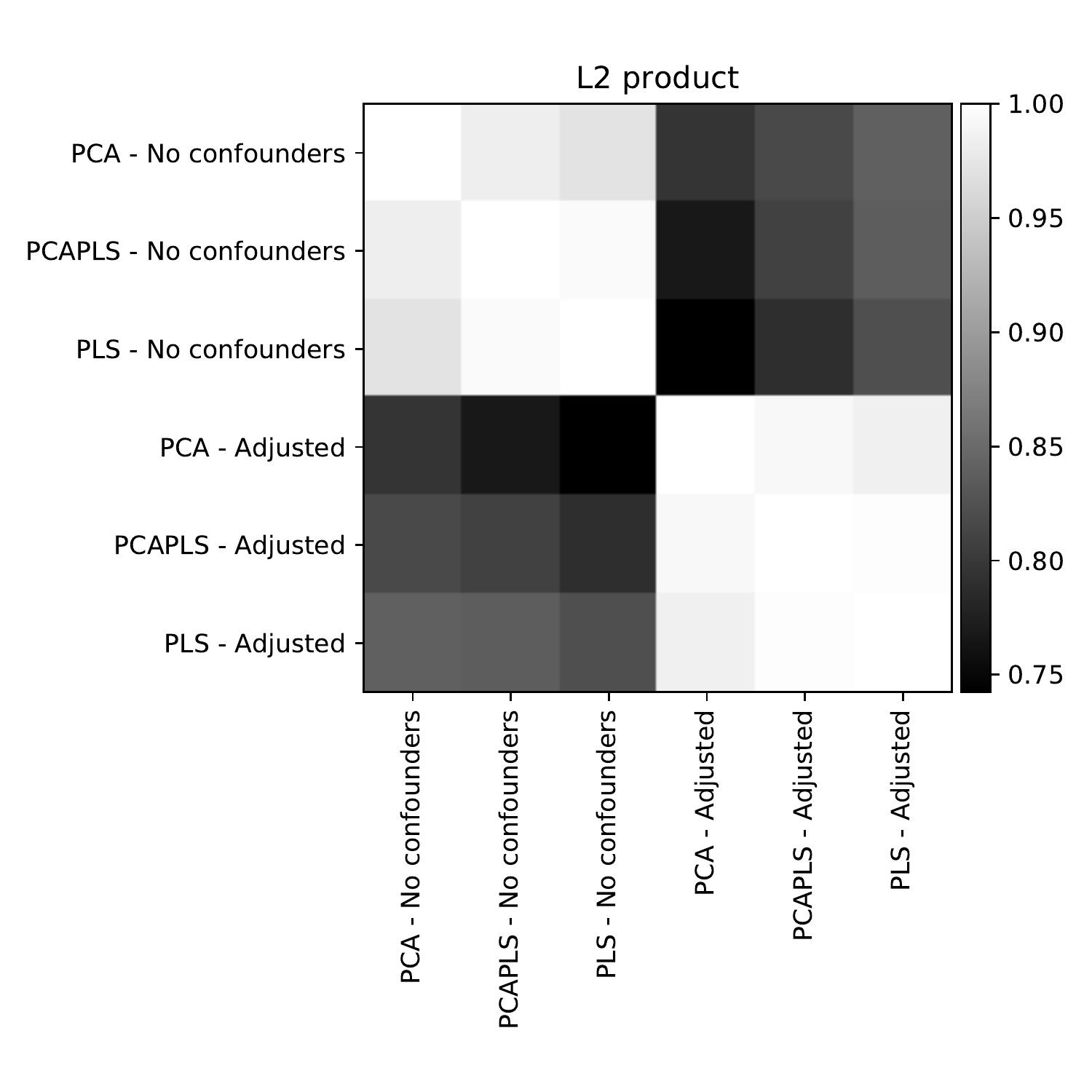}
\caption{Original shapes.}
\label{methodology:figure:differentModelsCorrelation}
\end{subfigure}
\begin{subfigure}[b]{0.4\textwidth}
\centering
\includegraphics[width = .9\textwidth]{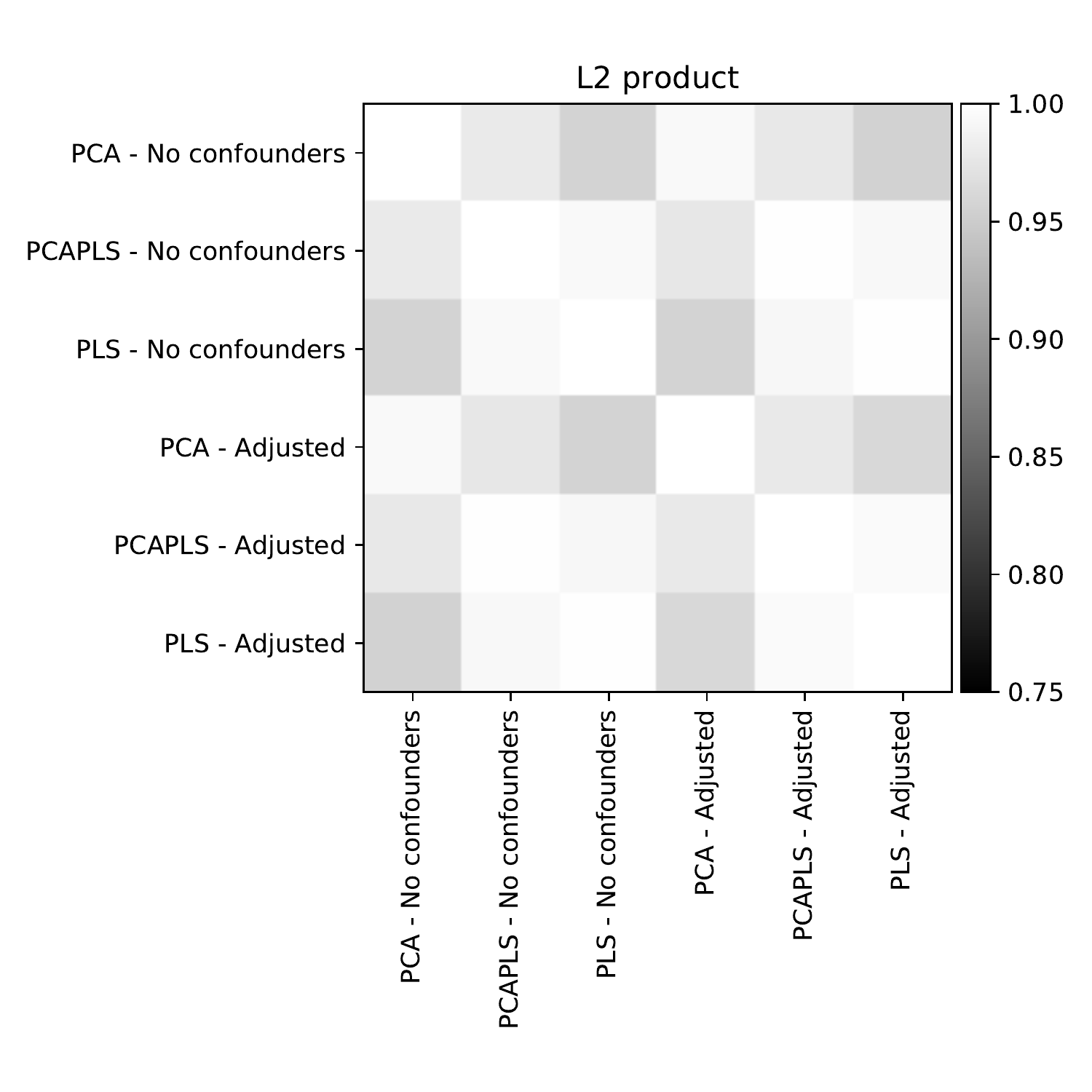}
\caption{Confounding deflation.}
\label{experiments:figure:correlationResidualModels}
\end{subfigure}
\caption[Comparison of different DR methods and confounding adjustment.]{$L_2$ product between the most discriminative shape pattern(the $L_2$ product between the most discriminative shape coincides with the correlation between scores) obtained using different combinations of \ac{DR} techniques and using or not the confounding adjustment in the logistic model. In subfigure \ref{methodology:figure:differentModelsCorrelation}, we can see that differences due to the \ac{DR} method were secondary to the adjustment when the shapes are not deflated. In subfigure \ref{experiments:figure:correlationResidualModels}, we can see that the differences due to adjustment disappear after the shapes are corrected using confounding deflation.}
\label{experiments:figure:correlationAllModels}
\end{figure*}

We applied our framework to identify the athletic remodelling in our dataset, and compared the effect of using the different \ac{DR} method and confounding-bias correction methods (confounding deflation and adjustment). We used the \ac{DR} parameters found in the previous section via \ac{CV}.  Figure \ref{experiments:figure:correlationAllModels} shows the $L_2$ product between all combinations of \ac{DR} methods and the adjustment or not by confounding variables. For the original (no confounding deflated) shapes (left), we found big differences due the inclusion or not of the confounding variables. The \ac{DR} choice gave only minor differences in this dataset: the resulting most discriminating modes of the \ac{DR} were very similar.  When the confounding deflation was applied (Figure \ref{experiments:figure:correlationResidualModels}), the use of the confounding adjustment became redundant and did not influence the resulting shape.

\begin{figure*}[!t]
\centering
\includegraphics[width = .8\textwidth]{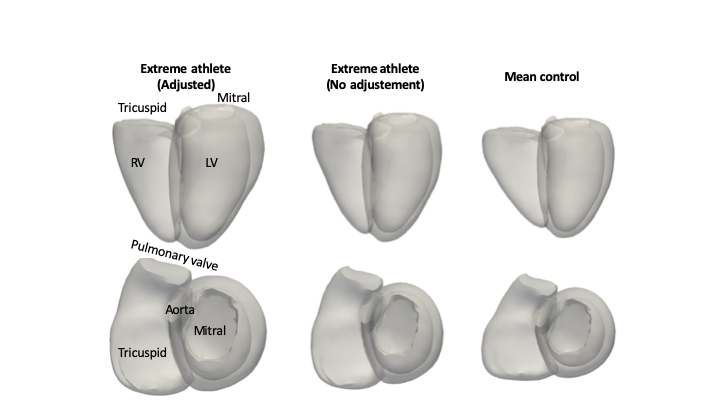}
\caption[Most discriminating shape patterns]{The picture depicts the mean shape  of the population (right), and that mean shape after applying  2STD of the athletic remodelling obtained after adjusting by confounding (left) and the same remodelling without the adjustment (center). Both remodelling patterns show a dilation of the ventricles, with a bigger dilation of the \ac{RV} outlet, but the one adjusted by confounders has a more pronounced dilation, and also a clear increase of the basal \ac{LV} wall thickness. The different rows correspond to two different views: the top row depicts a longitudinal view of the anterior wall of both ventricles and the bottom corresponds to a short axis view of the base, with the observer located in the atria.}
\label{methodology:figure:fullPopulationPCA}
\end{figure*}

\begin{figure*}[!t]
\centering
\includegraphics[width = .8\textwidth]{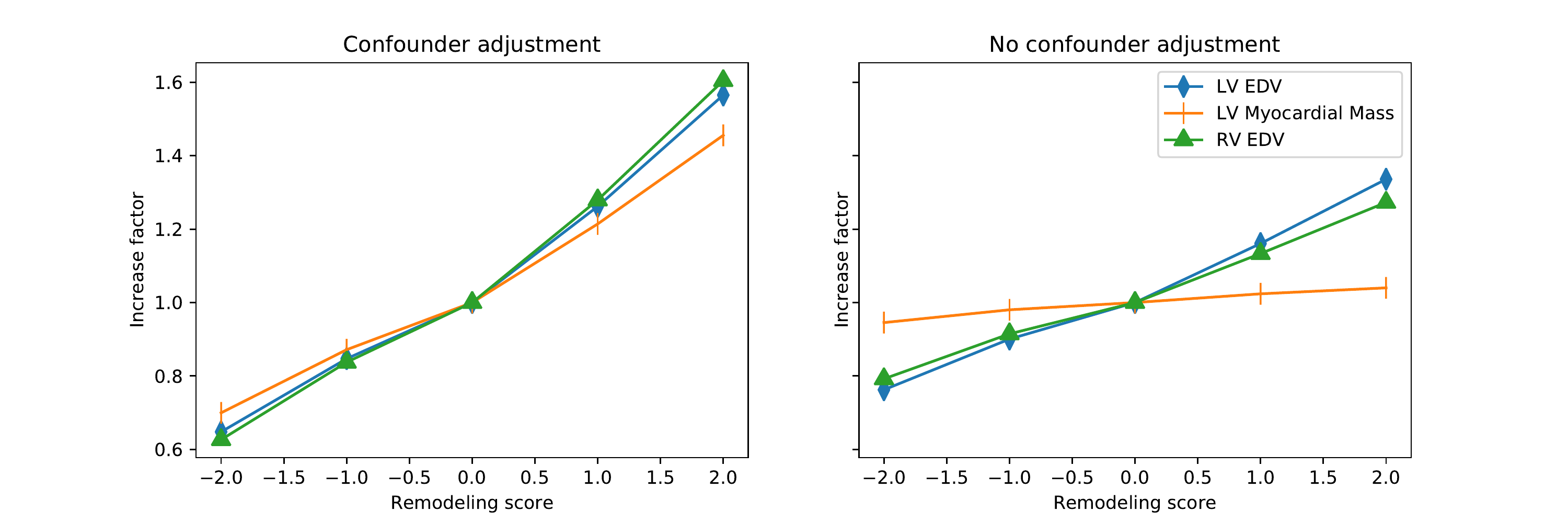}
\caption[Effect of adjustment in the measurement response]{The plot shows the measurement response to the remodelling: for each of the synthetic meshes generated by adding the shape remodelling pattern to the mean shape with different magnitude, we compute classical measurements and show the variation ratio with respect to the mean shape.}
\label{methodology:figure:fullPopulationMeasurements}
\end{figure*}

Figure \ref{methodology:figure:fullPopulationPCA} depicts the most discriminative shape patterns, expressed as mean + 2 STD,  obtained by the \ac{PCA} model with and without adjustment. Figure \ref{methodology:figure:fullPopulationMeasurements} shows their asociated measurement response. Both the adjusted and unadjusted models found an increase of the ventricular volumes that is the same for both \ac{LV} and \ac{RV}. Both models found shape changes in the \ac{RV}: the outflow region dilated more than the inlet base and apex.  Regarding myocardial mass, however, the models gave different results.  After confounder adjustment, we observed a large increase of \ac{LV} myocardial mass similar to ventricular dilation, while the unadjusted model found no difference in the mass.  The increase of mass of the adjusted model is concentrated in the base and lower in the apex: the base of the \ac{LV} is flatter and is exposed to bigger mechanical stress, so it is more prone to compensatory myocardial thickening. Finally, the increase of ventricular volume in the adjusted model is more pronounced than in the unadjusted model. 

\begin{figure*}[!t]
\centering
\includegraphics[width = .8 \textwidth]{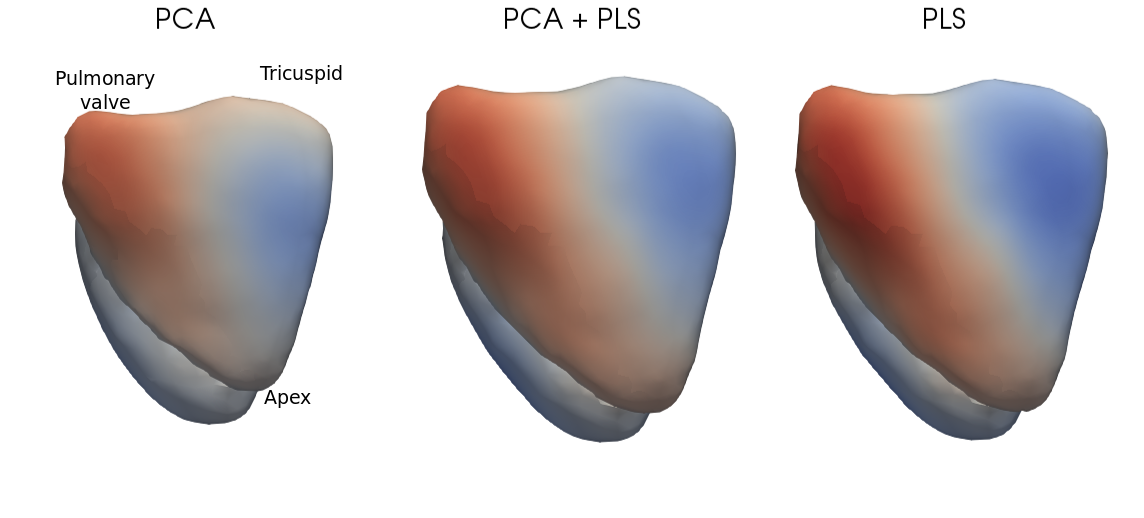}
\caption[Close-up on the right ventricular free wall of the different models predicting the athletic remodelling shape pattern]{Close-up on the right ventricular free wall of the different models predicting the athletic remodelling shape pattern. The red-blue color map encodes the regional amount of remodelling: red means big differences compared to the control population, and blue small/ no remodelling. We can see that \ac{PCA} remodelling is smoothly distributed through the whole ventricle, while the results obtained with \ac{PLS} and \ac{PCA} + \ac{PLS}  present a  more localized remodelling with a sharper red-blue transition: the remodelling is concentrated in the outflow.}
\label{methodology:figure:differenceDRModels}
\end{figure*}

The adjustment or not by confounders resulted in big differences between the remodelling patterns found. On the other hand, differences due to \ac{DR} were smaller, yet noteworthy. Figure \ref{methodology:figure:differenceDRModels} shows the remodelling patterns found for the different \ac{DR} techniques, with a colormap showing the local amount of remodelling: reed indicates substantial changes and blue no remodelling. The figure shows a view of the \ac{RV} free wall, which is the region that experienced more shape changes.  All 3 shape patterns were similar and followed the same global trends, but they presented regional differences. We observed that both \ac{PLS} and \ac{PCA} + \ac{PLS}  found a remodelling that was localised in the \ac{RV} outlet and , at a lower degree, in the apex, while \ac{PCA} showed a more spatially distributed remodelling that also affected the base and had a smoother transition between the affected and unaffected areas.
 
\subsection{BMI model}
\label{experiments:regressionBMI}
To have a better understanding of the effect of an elevated \ac{BMI} in the ventricles, we constructed a regression model that predicts \ac{BMI} from the cardiac shape. As athletes and controls had different ranges of \ac{BMI}, and to avoid finding any interference with the remodelling due to endurance sport, we only use the controls to build the \ac{BMI} model. Also, since \ac{BMI} and body size are related, we did not use the confounding variables in this model. The model was built using a \ac{PLS} in its original regression mode, using $3$ dimensions. To evaluate the model prediction capability, we computed the determination coefficient $R^2$ using 5-fold \ac{CV}, which was $0.44$. Similar to classification, we obtained synthetic representative shapes ($\hat{x}_b$) associated with certain \ac{BMI} value $b$. The synthetic mesh associated with a certain \ac{BMI} was the one having minimal-distance to the mean shape (using the Mahalanobis distance), constrained to having the required predicted \ac{BMI}. 

\begin{figure*}[!t]
\centering
\includegraphics[width = .6\textwidth]{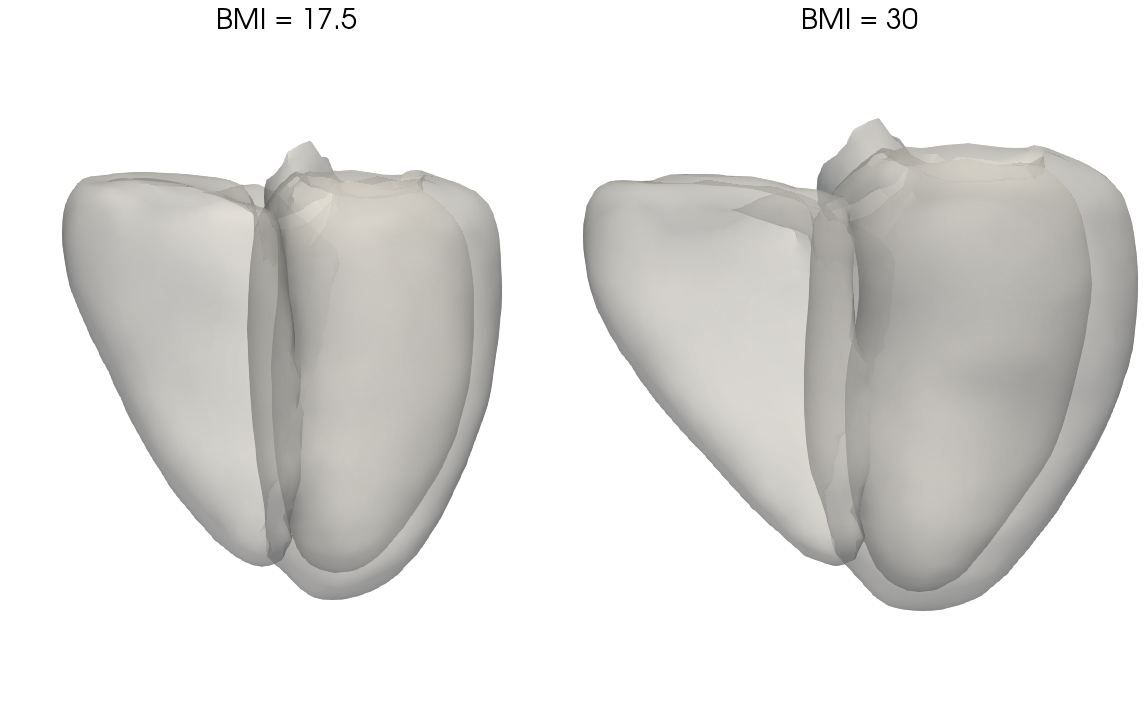}
\caption[Representative shapes of the BMI model]{Synthetic representative shapes of a patient with \ac{BMI} of 17.5 and another with \ac{BMI} 30 according to the \ac{BMI} predicting model. The remodelling consisted of an increase of volume (specially in the  axial directions) and \ac{LV} myocardial mass.}
\label{experiments:figure:bmiModel}
\end{figure*}
\begin{figure*}[!t]
\centering
\includegraphics[width = .4\textwidth]{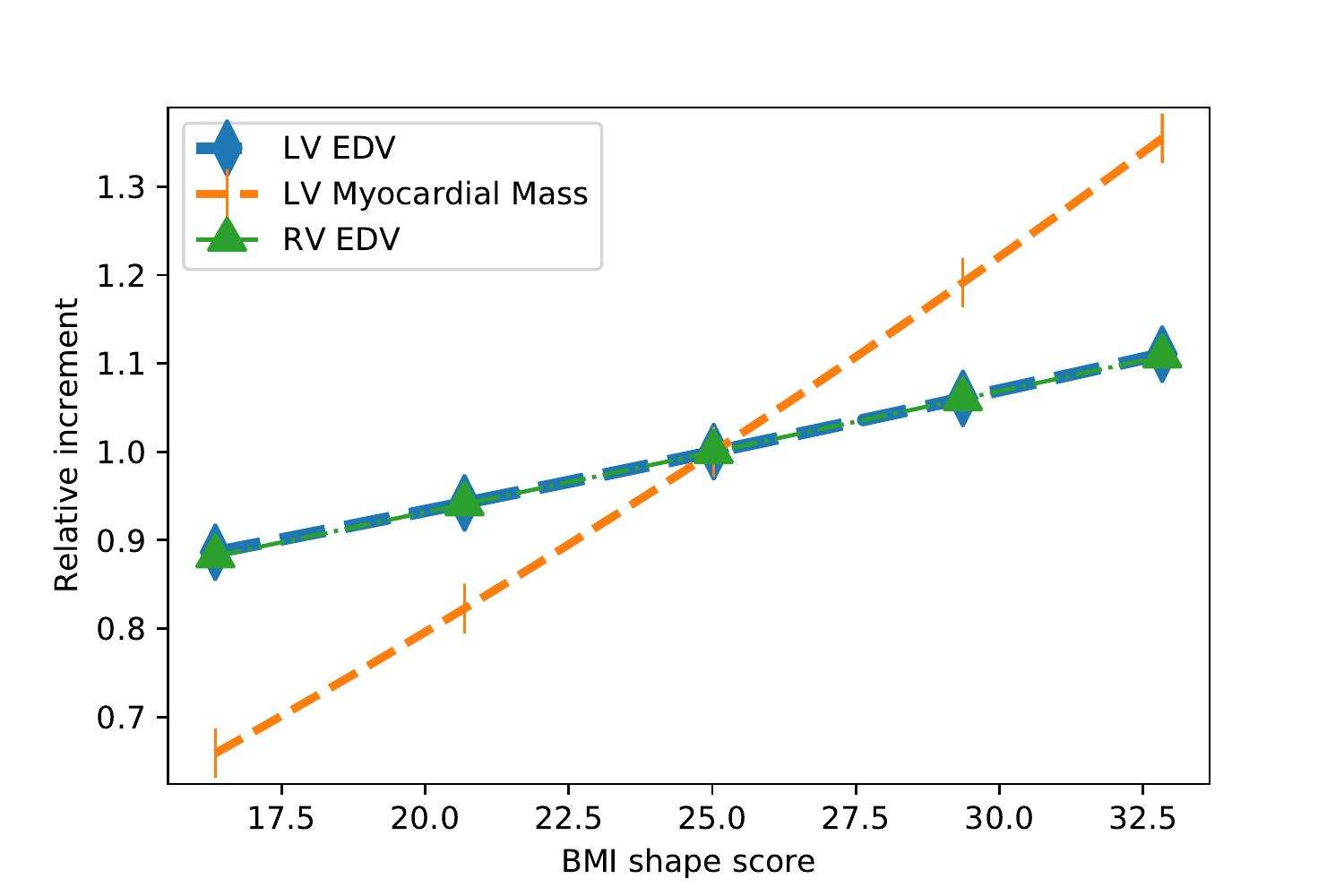}
\caption[Measurement response of the BMI-related shape changes.]{Measurement response of the BMI-related shape changes. As seen in the visual representation of the shapes, its main component is an increase of the myocardial mass, complemented with a smaller increase of volumes. }
\label{experiments:figure:bmiMeasurements}
\end{figure*}

The representative shapes associated to BMI values of 17.5 and 30, which are very extreme values, can be seen in Figure \ref{experiments:figure:bmiModel}, and Figure \ref{experiments:figure:bmiMeasurements} shows the measurement response of the remodelling. The \ac{BMI}-associated remodelling consists of a moderate increase of ventricular size and a bigger increment of the myocardial mass.

\subsection{Confounding adjustment}
\begin{figure*}[!t]
\centering
\begin{subfigure}[b]{0.45\textwidth}
\includegraphics[width = .99\textwidth]{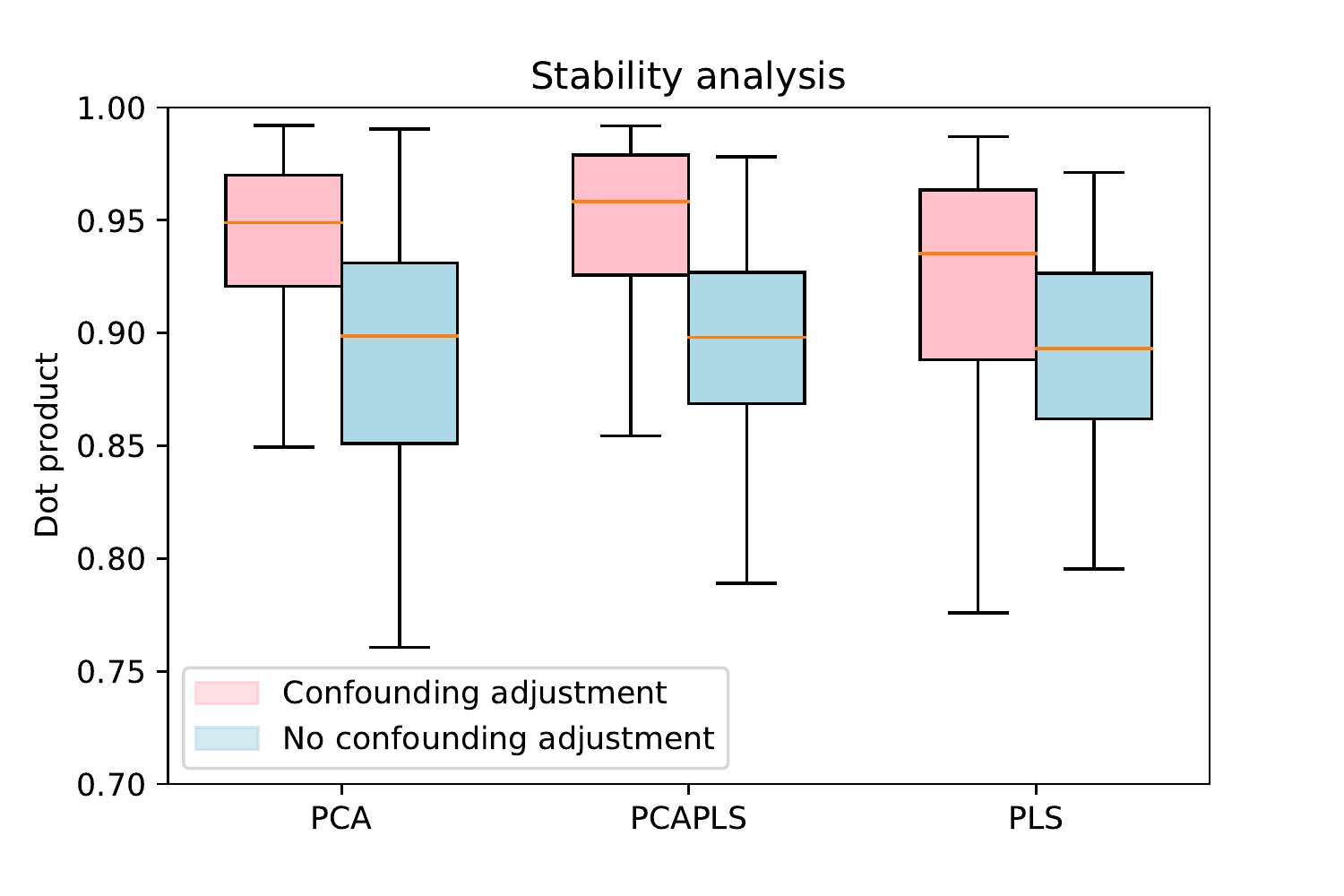}
\caption{Stability to imbalance}
\label{experiments:figure:stabilityRaw}
\end{subfigure}
\begin{subfigure}[b]{0.45\textwidth}
\includegraphics[width = .99\textwidth]{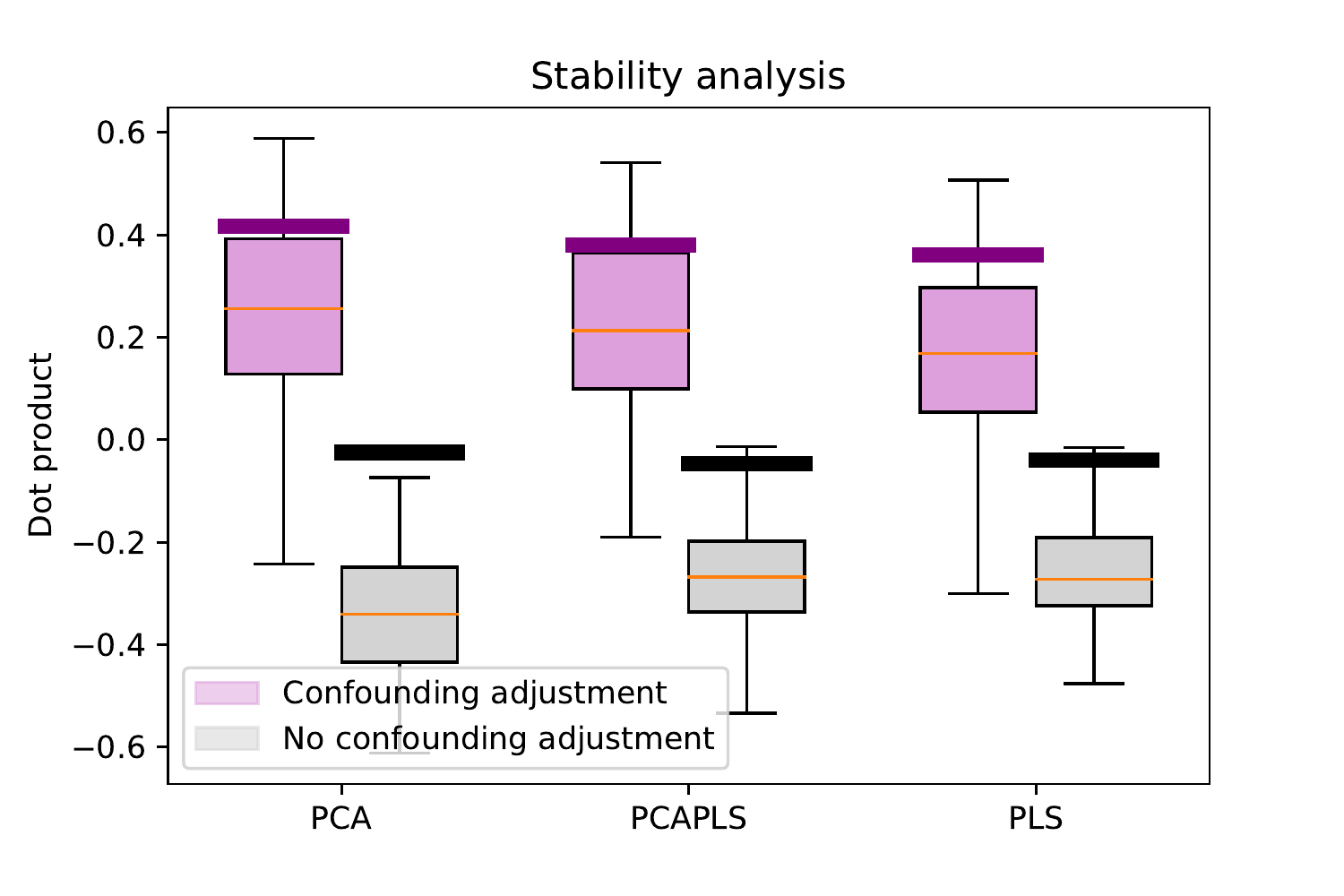}
\caption{Dot product with BMI model. }
\label{experiments:figure:bmiRaw}
\end{subfigure}
\caption[Stability of adjustment in BMI-imbalanced datasets.]{Effect of using confounding adjustment in the stability of the models derived from the imbalanced datasets for the 3 different DR methods. a) agreement of the model trained on downsampled data with the one on full data, measured via its dot product. b) The thick solid lines correspond to the dot product between the BMI shape remodelling pattern, and the athletic shape remodelling derived from the full population. The boxplots show the dot product between the downsampled derived shape patterns and the \ac{BMI} shape remodelling pattern. When adjustment is used, the athletic and \ac{BMI} mode have a positive relation (ie, they have remodelling partially in the same direction), due to both  partially responding to an increase of pressure; however, that relation can become negative after downsampling, since the controls become more overweight.}
\end{figure*}

\begin{figure*}[!t]
\centering
\begin{subfigure}[b]{0.8\textwidth}

\includegraphics[width = .99\textwidth]{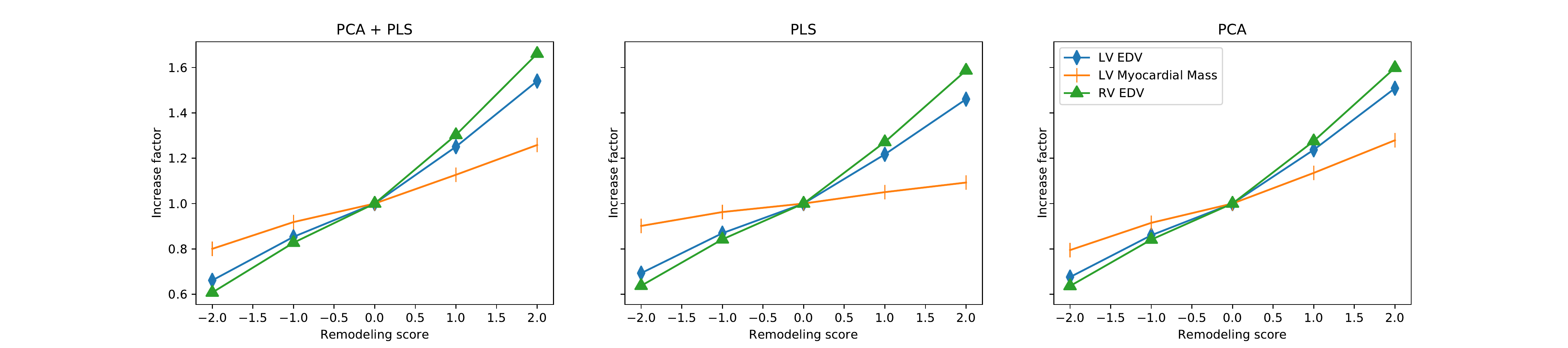}
\centering
\caption{Measurement response with confounder adjustment. }
\label{experiments:figure:measurementsConfRaw}
\end{subfigure}
\\
\begin{subfigure}[b]{0.8\textwidth}
\centering
\includegraphics[width = .99\textwidth]{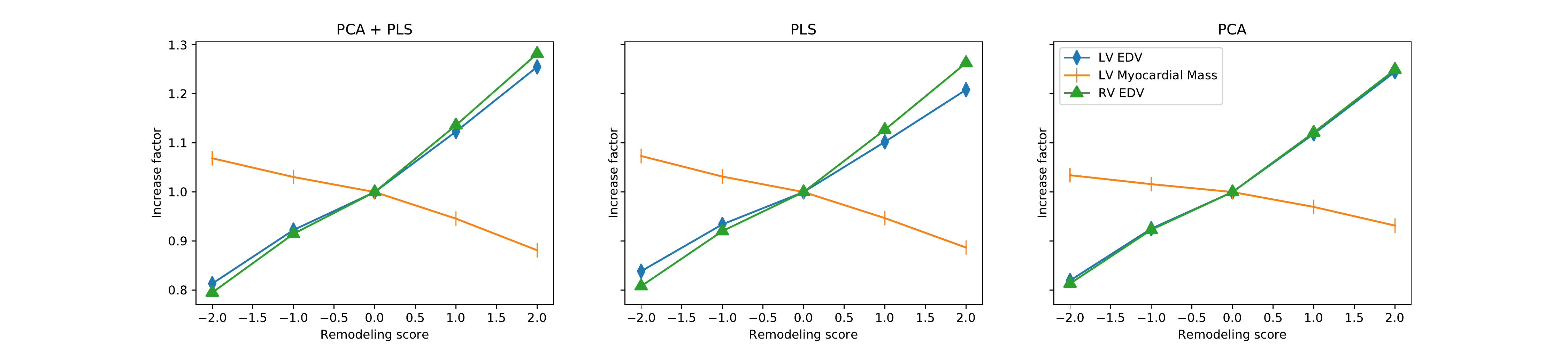}
\caption{Measurement response without confounder adjustment. }
\label{experiments:figure:measurementsNoConfRaw}
\end{subfigure}
\caption[Measurement response of the models trained on the downsampled population]{Measurement response of the downsampled population with and without adjustment. We can see that the unadjusted methods find a negative relationship between athletic remodelling and \ac{LV} mass, but the adjusted methods find a positive relationship. Figure \ref{methodology:figure:fullPopulationMeasurements} shows the equivalent plots for the models trained with the full population, which we use as groundtruth. The adjusted models are more similar to the groundtruth than the unadjusted.}
\end{figure*}

Figure \ref{experiments:figure:stabilityRaw} depicts the $L_2$ products between the athletic remodelling shape pattern (normalised to be a unitary vector) derived from the full population, and the shape pattern derived from the \ac{BMI}-imbalanced populations. We found that without adjustment the imbalance confused the method, and it mixed the differences due to \ac{BMI} with the ones related to endurance training. The adjustment by confounding variables resulted in a better agreement between the downsampled-derived remodelling patterns and the full-population pattern than the unadjusted model, as can be seen in Figure \ref{experiments:figure:measurementsConfRaw}. Figure \ref{experiments:figure:bmiRaw} shows the dot product between the most discriminating shape pattern obtained with the downsampled (boxplot) and full (thick solid lines) datasets, and also the shape pattern associated to \ac{BMI}, computed as described in section \ref{experiments:regressionBMI}. There, we can see that in the downsampled datasets, the \ac{BMI}-associated remodelling and the athletic one had lower correlation. This drop of correlation was larger for the models unadjusted by confounder variables.

\begin{figure*}[!t]
\centering
\includegraphics[width = .8\textwidth]{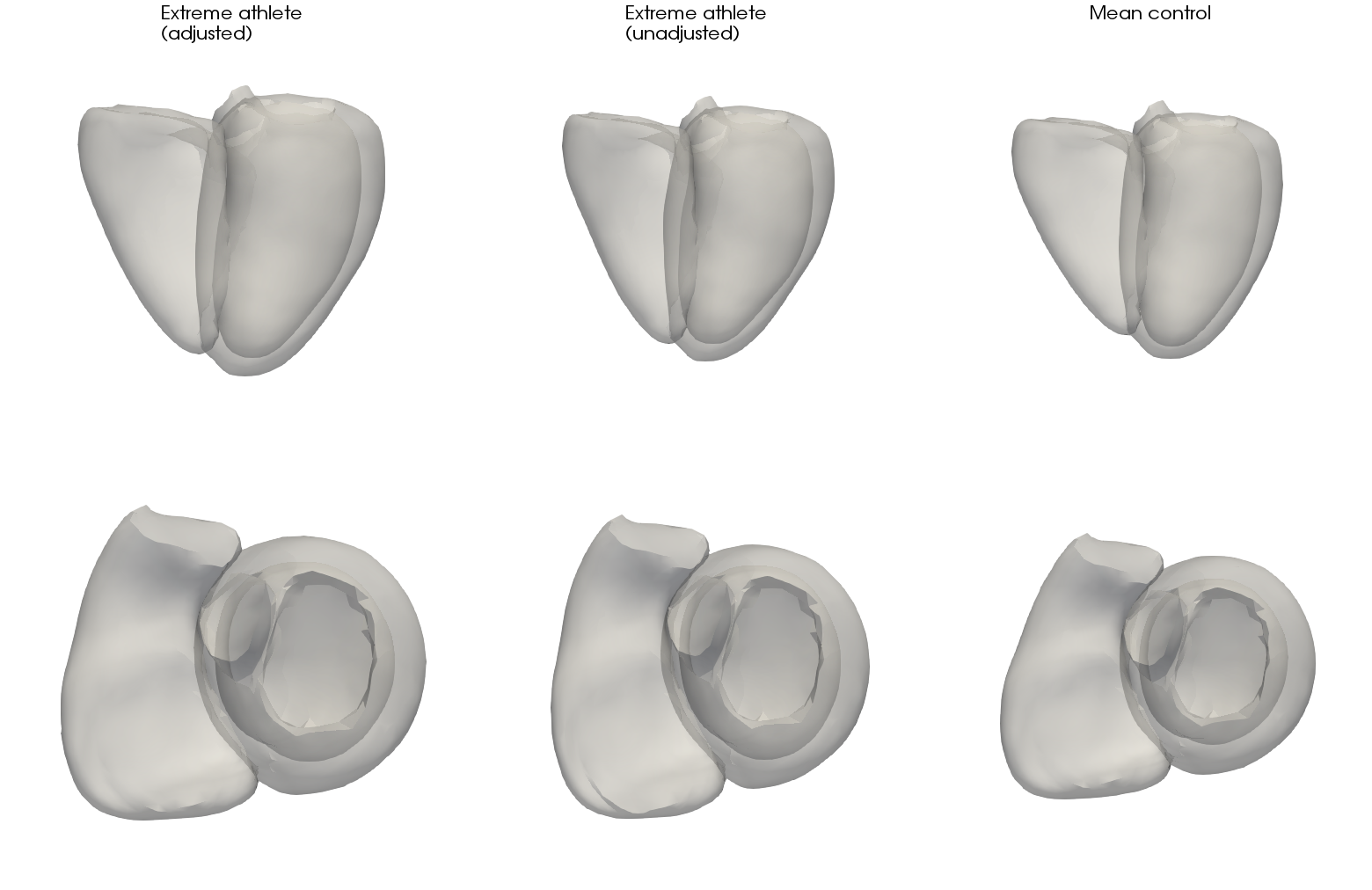}
\caption[Representative shapes of models trained on the downsampled population]{Mean shape (right), and mean shape + 2STD  of the athletic shape remodelling pattern(left), derived from a BMI-imbalanced population. The pattern in the center column was obtained without adjustment, while the one on the right was adjusted. We can that the unadjusted model finds a decrease of myocardial mass in the apical and septal walls, while the adjusted finds an increase of mass}
\label{experiments:figure:athletesModelsDownsampled}
\end{figure*}

Figure \ref{experiments:figure:measurementsConfRaw} shows the measurement response to the athletic remodelling of several models trained with a downsampled population. It shows different \ac{DR} methods with confounding adjustment  (upper row) and without confounding adjustment (lower row). We can see that when no confounding correction is used, the \textcolor{black}{remodelling} found is associated with lower myocardial mass, contrary to what is observed with the full dataset experiment in which myocardial mass was maintained. When the shape is adjusted by confounders, athletic remodelling was again associated with an increase of myocardial mass, consistent with the full population model. Figure \ref{experiments:figure:athletesModelsDownsampled} shows the discriminating shape patterns for a randomly selected downsampled dataset. There, we can observe that the model associates athletic remodelling with a decrease of the wall thickness in the septum and apical regions of the \ac{LV} when confounders are not considered.

Resulting shape patterns were affected by the \ac{DR} methods, \ac{PLS} was more unstable than \ac{PCA} and \ac{PCA} + \ac{PLS}, who had a better correlation to their full-population discriminating shape. In Figure \ref{experiments:figure:measurementsConfRaw}, we can see that \ac{PLS} was not able to recover the increase of myocardial mass, and its mode presented no variation in the myocardial mass while both \ac{PCA} and \ac{PCA} + \ac{PLS} could.

\subsection{Confounding deflation}

In this subsection we analyze the effect of the confounding deflation. As stated above, the shape prediction model of the confounding deflation was trained using only the controls. We tested two possible scenarios: the case in which the athletes population was downsampled, and therefore the building of the shape residual model was not affected; and when the controls were downsampled (as in the previous subsection). The former was the most appropriate situation to apply confounding deflation, since there would not be extra unstability/bias introduced during the  confounding deflation step, while the latter could introduce the bias in the dataset during the confounding deflation process. Finally, we show the potential danger of training the residual model using both populations and how the confounding deflation could even increment bias. 

\begin{figure*}[!t]
\centering
\begin{subfigure}[b]{0.4\textwidth}
\includegraphics[width = .95 \textwidth]{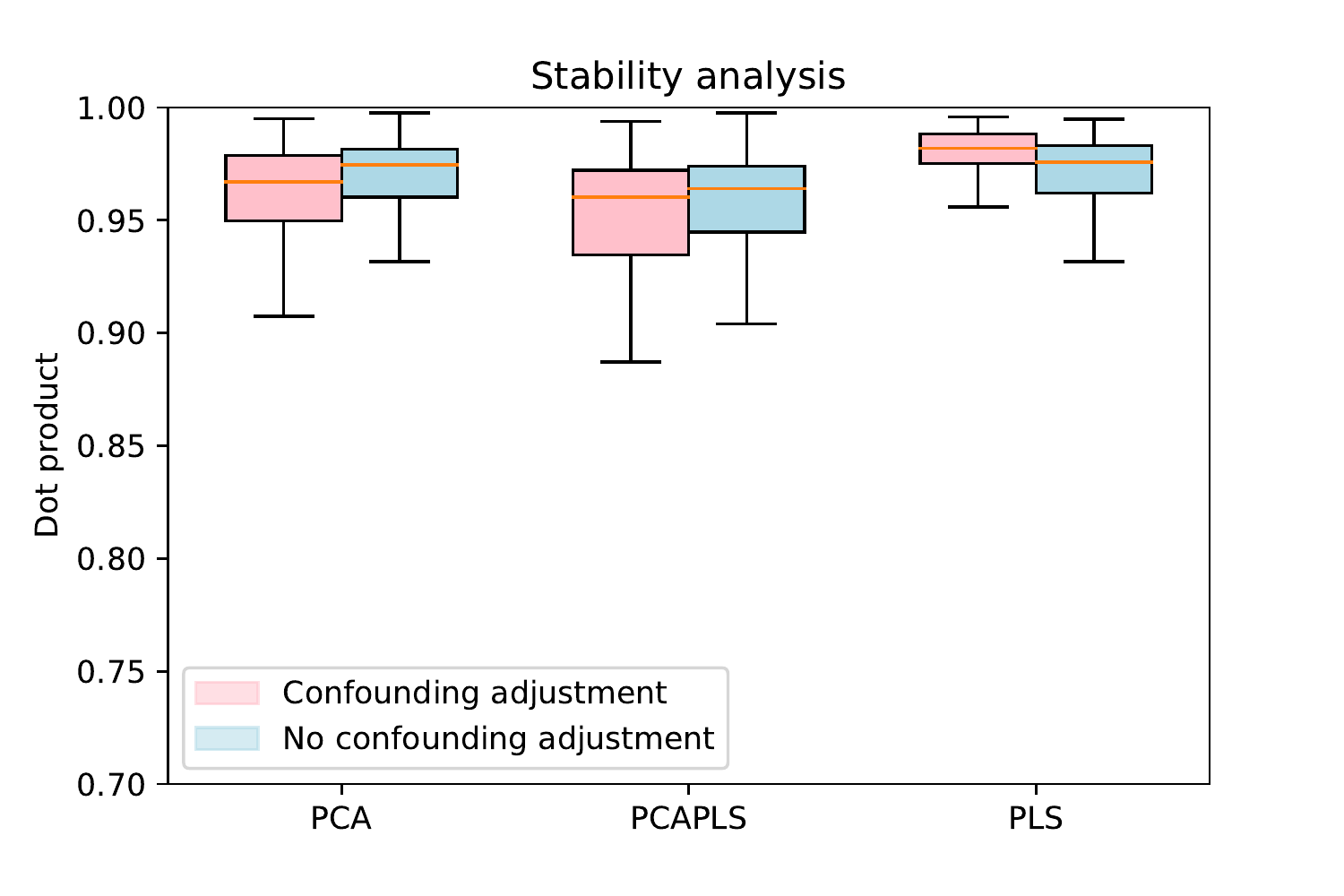}
\caption{Dot product with full data model. }
\label{experiments:figure:stabilityRes}
\end{subfigure}
\begin{subfigure}[b]{0.4\textwidth}
\centering
\includegraphics[width = .95 \textwidth]{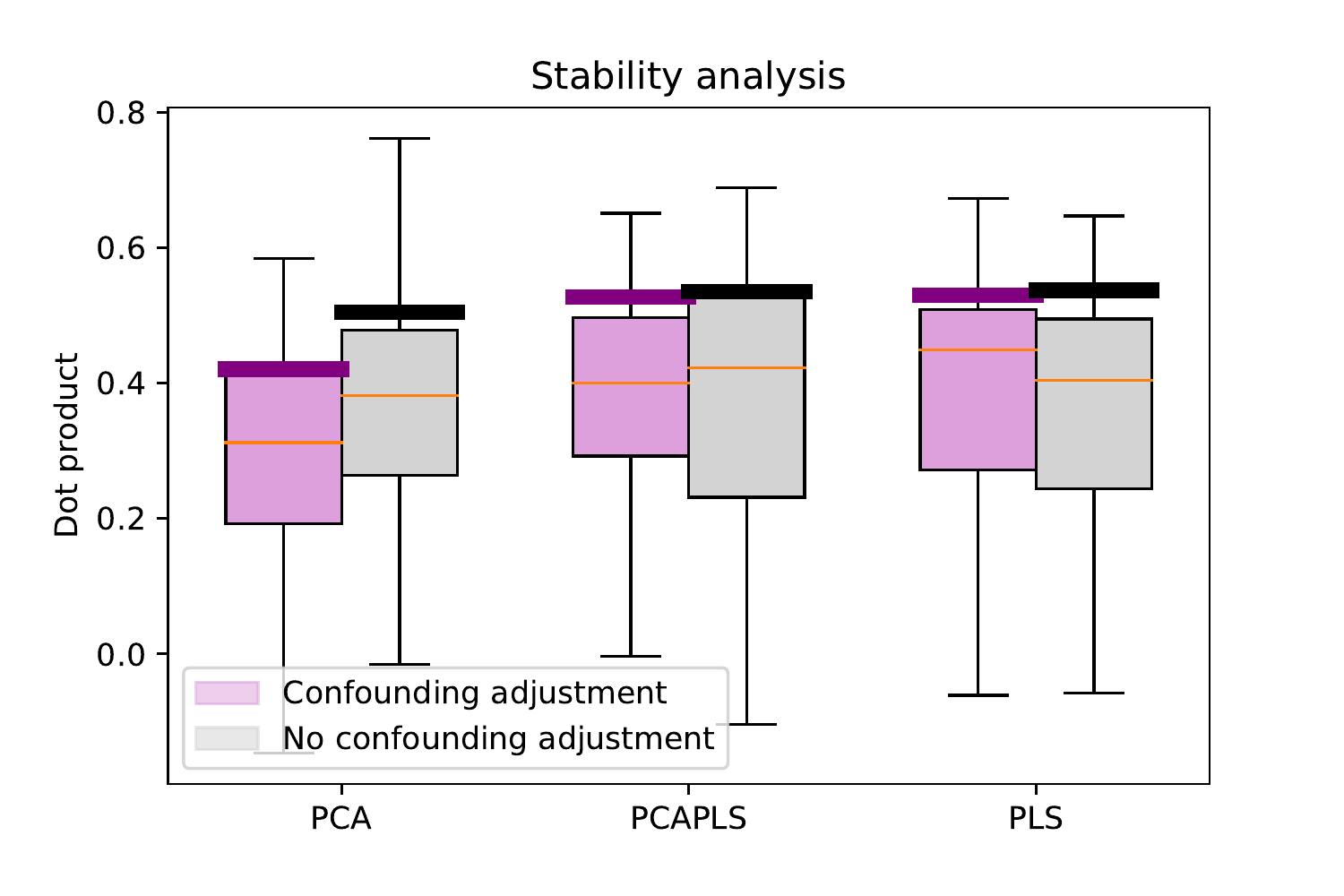}
\caption{Dot product to \ac{BMI} model for residual. }
\label{experiments:figure:bmiRes}
\end{subfigure}
\caption[Stability analysis of the confounding deflation method trained with a non-downsampled population.]{Stability analysis of the confounding deflation method when the athletes (that are not used in the construction of the residual model) are downsampled. }
\end{figure*}

When controls were not downsampled, the population used to train the shape prediction model was relatively unbiased. Instead of downsampling the controls, we downsampled the athletes, removing athletes with high \ac{BMI} analogously to the procedure used to downsample the controls. Figure \ref{experiments:figure:stabilityRes} and \ref{experiments:figure:bmiRes} depict the same experiments as in the previous section: the dot product between the most discriminating shape obtained with the downsampled data and the remodelling obtained considering the whole population, and also the dot product with the \ac{BMI} mode. Results showed a considerable decrease in variability and a higher correlation with the full-dataset-derived remodelling than the  confounder adjustment experiments (Figure \ref{experiments:figure:stabilityRaw}). Adding confounder adjustment on top of confounding deflation did not produce an increase of accuracy.

\begin{figure*}[!t]
\centering
\begin{subfigure}[b]{0.4\textwidth}

\includegraphics[width = .95 \textwidth]{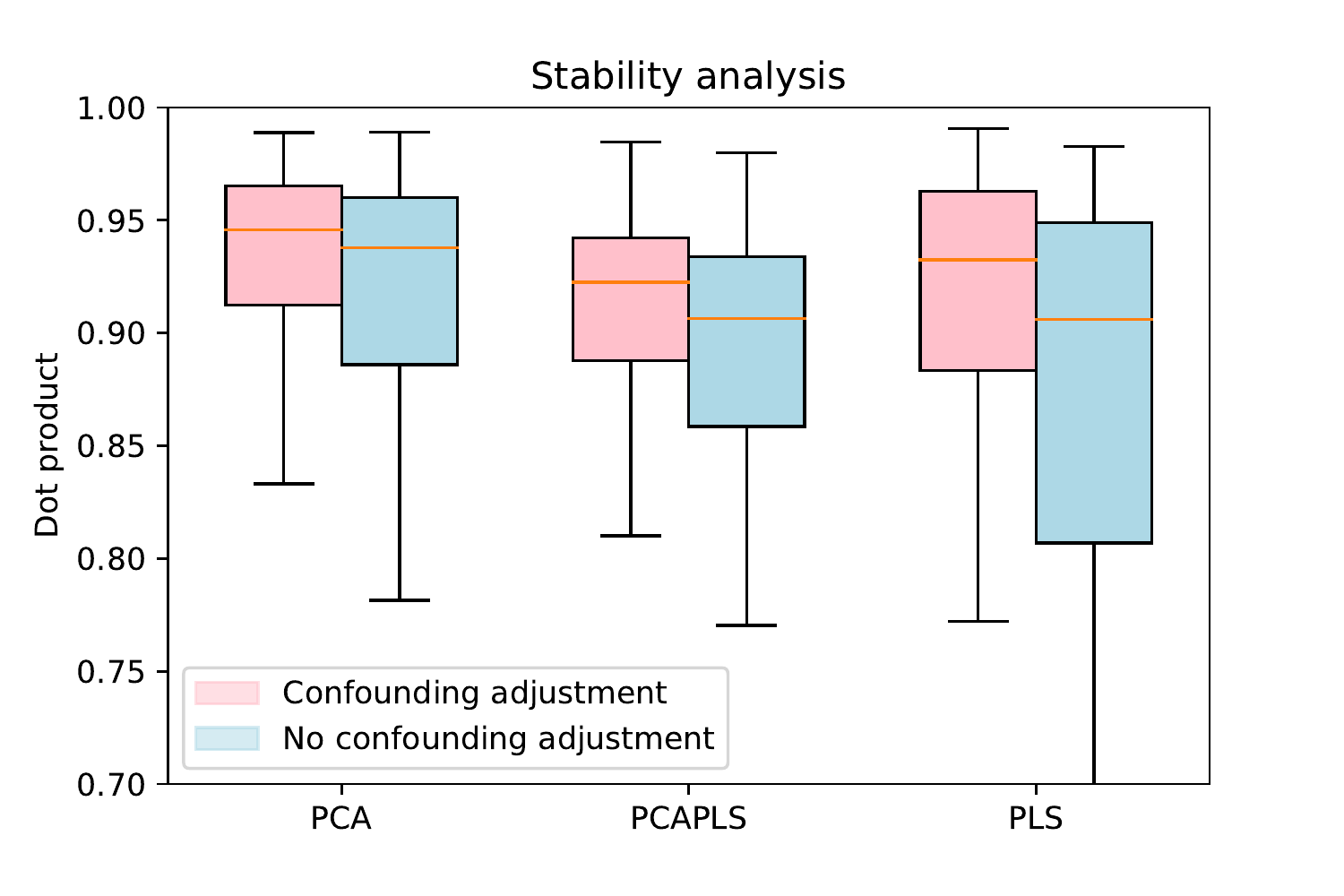}
\caption{Confounder deflation trained on controls only.}
\label{experiments:figure:stabilityResDownControls}
\end{subfigure}
\begin{subfigure}[b]{0.4\textwidth}
\includegraphics[width = .95 \textwidth]{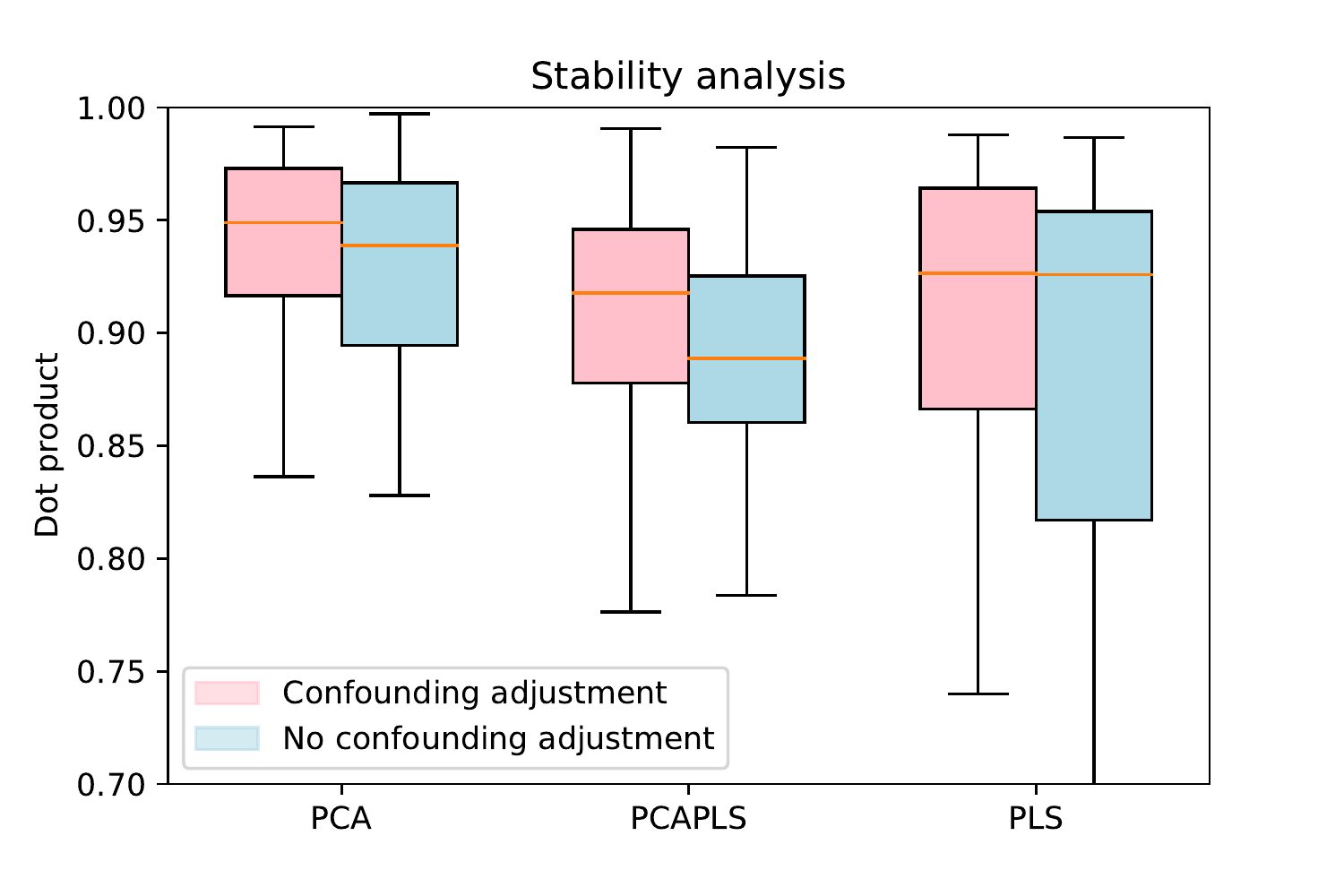}
\caption{Confounder deflation trained on both classes.}
\label{experiments:figure:stabilityResDownsampledBothPopulations}
\end{subfigure}
\caption[Effect of the population used to train the confounder deflation model on the discriminative pattern stability.] {Effect of the population used to train the confounder deflation model on the discriminative pattern stability, assessed via its $L_2$ product with the result obtained using the full population. Subfigure a) shows when the training of the shape prediction model in the confounder deflation step is trained using the downsampled class, and subfigure b) shows when both controls and athletes are used for training.}
\end{figure*}

Figure \ref{experiments:figure:stabilityResDownControls} shows the results when the population in which the residual is trained is downsampled. We downsampled the controls based on their \ac{BMI}. Results were  much worse than when athletes were downsampled, and  even worse than a simple confounder adjustment. Adding confounding adjustment  on top of confounding deflation had a beneficial effect. Figure \ref{experiments:figure:stabilityResDownsampledBothPopulations} shows the effect of using both athletes and controls in the training of the shape-prediction model for confounding deflation: there was a drop in stability compared to the use of a single population when \ac{PLS} and \ac{PCA} + \ac{PLS} were used (Figure \ref{experiments:figure:stabilityRaw}). Strangely there was an improvement compared to the baseline (where confounders are completely ignored).

\begin{figure*}[!t]
\centering
\includegraphics[width = .65 \textwidth]{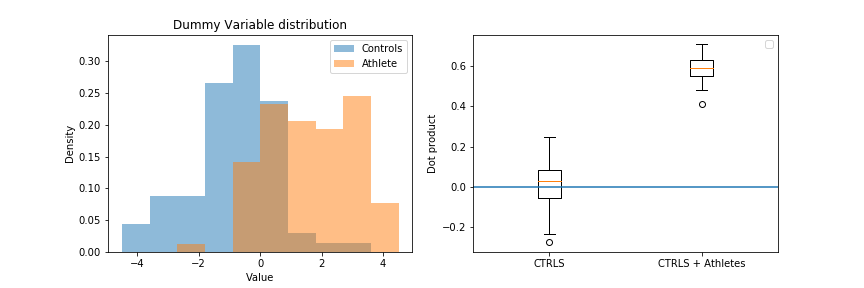}
\caption[Distribution of the dummy variable and its dot product with the athletic remodelling shape pattern]{This figure shows the distribution of the dummy variable used as a confounder,  and the dot product between the regression coefficients associated  to the dummy variable in the residual model and the athletic remodelling shape pattern. }
\label{experiments:figure:artificialData}
\end{figure*}

Therefore we observed that using both populations in the training of the shape prediction model during the confounding deflation step resulted in worse results. We explored the reason why using both populations can create a confounding effect, associating an imbalance in a variable to the inter-class shape differences. This can happen even when the variable is not associated to any shape remodelling. To illustrate this effect, we created a dummy synthetic variable which is just the athlete label plus Gaussian noise. Obviously, by generation we knew that this variable did not have any direct relationship to shape. To evaluate this effect, we computed the $L_2$ dot product with the most discriminating shape between athletes and controls and the shape pattern associated to the dummy variable in the shape-prediction model.  We repeated this full variable generation and $L_2$ products computation process $100$ times to remove randomness of the analysis. Figure \ref{experiments:figure:artificialData} (left) shows the distribution of this dummy variable for a certain seed. We  constructed two residual models one only with the controls, and the other with both populations. Figure  \ref{experiments:figure:artificialData} (right) shows the distribution of the dot product of the shape prediction model coefficients associated to the dummy variable when both populations are used, and when only the controls are used.  The shape associated to the dummy variable is independent to the athletic remodelling when only one population is used, but becomes very similar to the control-athlete difference when both populations are used in training due to confounding effect.

\section{Discussion}
The shape models corresponding to the athletic and overweight remodelling  corroborated the current clinical literature. The \ac{BMI} remodelling consisted in mostly a concentric remodelling, both the \ac{LV} mass and \ac{LV} EDV increased, increasing more the mass than the volume: this coincides with the hypothesis that remodelling is concentric to cope with elevated pressure (increase of myocardial mass) and \ac{CO} demands (increase of volume).  Another hint that the remodelling was mostly pressure driven is that the septum  of the mesh representing a high \ac{BMI} individual presents a bulge below the aorta, that has been described as an early indicator of elevated pressure \citep{Baltabaeva2008,Gaudron2016}. The athletic remodelling consisted of a predominant increase of volume, and if confounder adjustment was performed, an increase of myocardial mass. There is some controversy if endurance athletes remodelling is more eccentric ( they increase their volume more than their mass) or concentric (they increase their mass more than their volume), but there is a consensus that there is both an increase of mass and volume \citep{Scharhag2002}, and therefore we considered the adjusted model more accurate. The increase of ventricular volume in the confounder adjusted model is considerably larger than the unadjusted. These differences between the adjusted and not adjusted models can be explained because the variability in size and myocardial mass can come also from the patients morphometrics: big persons have big hearts. Without considering this extra non-imaging information, it is not possible to discern when the size of the heart is due to the patient being big, or the heart had a dilation as a remodelling reaction to exercise. 

The artificially generated imbalanced sets allowed us to validate our hypothesis: without any correction, the athletic remodelling presented a reduction in \ac{LV} myocardial mass. This is obviously false, since all studies have found that endurance exercise provokes an increase of myocardial mass. The downsampled control population had a high percentage of overweight, who had a concentric remodelling and causing the previous association with controls and higher myocardial mass in the unadjusted model. The stability analysis, also showed a bigger match between the downsampled-derived remodelling and the full-dataset one when confounding adjustment was used. 

We also studied another strategy: confounding deflation. Confounding deflation consists of generating a model that predicts shape from the confounders, and working with the residual of that prediction. Our results showed that this strategy worked well when there was access to a good population to train the shape prediction model: otherwise it can actually increase bias. This was shown when training on the downsampled population, or when both the case and control were used simultaneously.

Finally, we compared different linear methods for \ac{DR}: \ac{PCA} and  \ac{PLS}. In the original dataset, we obtained a better classification accuracy using \ac{PLS}. We also showed that \ac{PLS} was able to capture more localised remodelling than \ac{PCA}, that was limited to smooth and global remodelling. However, \ac{PCA} \textcolor{black}{outperformed} \ac{PLS} in stability tested during the downsampling analysis. Both models could be combined by using a coarse \ac{PCA}, that only removes the modes encoding very little variance,  with a \ac{PLS}. With this method we were able to capture both localised remodelling and have high stability to imbalances. \textcolor{black}{In our work, we have focused on linear \ac{DR} methods, but the same results are obtained using non-linear methods: an experiment using U-MAP can be found in supplementary material S2 and we observed the same confounding effect that was partially corrected after using confounding adjustment.}

\textcolor{black}{A limitation of our work is that only shape features were considered. It remains to be seen if intensity-based features, such as radiomics \citep{Cetin2018ACine-MRI}, can provide added value to the classification problem. Another limitation is that only 
\ac{ED} was analysed using this technique, which might be enough for studying athletic remodelling. For other conditions that  primarily affect the systolic phase, the full cardiac dynamics should be analysed using spatio-temporal atlases.  Furthermore, since only \ac{SA} images were used for constructing the model, the apical region could not be analysed accurately: for assessing apical remodelling, 3D-echocardiography might be better suited.}

\section{Conclusion}
We have presented a \ac{SSA} framework to find regional shape differences between two populations, taking special care to correct for any potential bias related to demographics parameters. The framework is fully linear: it used a \ac{PCA} sequentially combined with a \ac{PLS} as dimensionality reduction of the shapes, followed by a logistic regression model. The linearity allows to easily interpret and visualise the model and build synthetic representative shapes of the model. To correct for confounding effects, it incorporates adjustment, where the confounding variables are added to the logistic model and confounding deflation, which consists of building a regression model that predicts shape from confounding variables, and we used it to remove the shape variability  associated to these variables. 

We applied our framework to a real dataset consisting of athletes and controls to find the remodelling due to the practice of endurance exercise. Our results confirmed the current literature on endurance-sport remodelling in the \ac{LV}: ventricular dilation and increment of myocardial mass, specially in the basal area. In the \ac{RV}, we found that the volume increase was not homogeneous but concentrated in the outflow.  In the controls, we used an adaptation of our classification framework to regression to explore obesity remodelling and found it to be mainly an increase of myocardial mass.


In this population, we analysed the effect of confounders in a semi-synthetic dataset obtained by downsampling the control population non-uniformly, keeping individuals with high \ac{BMI}. Even if athletic remodelling is very prominent, we were able to bias the model to output that athletes have lower myocardial mass than controls. This was corrected when adjustment was used. However, we found that we could only use confounding deflation when the control population was relatively big and balanced, and if that is not the case using confounding deflation can actually increment bias. \textcolor{black}{Here, we presented an example of a cardiology application, but the proposed method is not unique for the heart. Future studies could identify which demographic variables influence the shape of other organs, based on the same methodology.}

In our work we have tested only linear \ac{SSA} methods, but this confounder-related problems might appear even more with the use of more complex frameworks, able to  capture non-linear shape patterns to capture subtler morphology differences in populations that are almost indistinguishable from controls and differences cannot be found through traditional means.

\section*{Acknowledgements}
We thank Dr Weese and Dr Groth from Philips Research Hamburg for the segmentation tool \textcolor{black}{and Dr Piella for fruitful discussions.}

\section*{Funding}
This study was partially supported by the Spanish Ministry of Economy and Competitiveness (grant DEP2013-44923-P, TIN2014-52923-R; Maria de Maeztu Units of Excellence Programme - MDM-2015-0502),  el Fondo Europeo de Desarrollo Regional  (FEDER) , the European Union under the Horizon 2020 Programme for Research, Innovation (grant agreement No. 642676 CardioFunXion) and  Erasmus+ Programme (Framework Agreement number: 2013-0040), “la Caixa” Foundation (LCF/PR/GN14/10270005, LCF/PR/GN18/10310003),  Instituto de Salud Carlos III (PI14/00226, PI17/00675)  integrated in the “Plan Nacional I+D+I” and AGAUR 2017 SGR grant nº 1531. 
\bibliography{libraryMendeley}
\end{document}